\documentclass[journal]{IEEEtran}

\IEEEoverridecommandlockouts          
\usepackage{booktabs}
\usepackage{subfigure}
\usepackage{graphics} 
\usepackage{epsfig}
\usepackage{graphicx} 
\usepackage{array}
\usepackage[flushleft]{threeparttable} 

\usepackage{xcolor, soul}
\usepackage{amssymb,amsmath,amsbsy} 
\usepackage{multirow}
\usepackage{multicol} 
\usepackage{hhline}
\usepackage{mathptmx} 

\begin{document}

\title{Video action recognition for lane-change classification and prediction of surrounding vehicles}

\author{Mahdi Biparva, 
        David Fern\'{a}ndez-Llorca,~\IEEEmembership{Senior,~IEEE,}
        Rub\'{e}n Izquierdo, 
        and~John~K.~Tsotsos,~\IEEEmembership{Felllow,~IEEE,}
\thanks{M. Biparva and J. K. Tsotsos are with the Department of Electrical Engineering and Computer Science, York University, Toronto, ON M3J 1P3, Canada (e-mail: mhdbprv@cse.yorku.ca; tsotsos@eecs.yorku.ca).}
\thanks{D. F. Llorca and R. Izquierdo are with the Computer Engineering Department, University of Alcal\'{a}, Alcal\'{a} de Henares, Madrid, Spain (e-mail: david.fernandezl@uah.es; ruben.izquierdo@uah.es). D. F. Llorca is also with the European Commission - Joint Research Center, Seville, Spain. }
\thanks{Manuscript received October XX, 2021; revised YYYY ZZ, 2022.}}

\markboth{IEEE Transactions on Intelligent Vehicles,~Vol.~XX, No.~YY, ZZZZ~2022}%
{Biparva \MakeLowercase{\textit{et al.}}: Action recognition for lane-change prediction of surrounding vehicles}


\maketitle

\begin{abstract}
In highway scenarios, an alert human driver will typically anticipate early cut-in/cut-out maneuvers of surrounding vehicles using visual cues mainly. Autonomous vehicles must anticipate these situations at an early stage too, to increase their safety and efficiency. In this work, lane-change recognition and prediction tasks are posed as video action recognition problems. Up to four different two-stream-based approaches, that have been successfully applied to address human action recognition, are adapted here by stacking visual cues from forward-looking video cameras to recognize and anticipate lane-changes of target vehicles. We study the influence of context and observation horizons on performance, and different prediction horizons are analyzed. The different models are trained and evaluated using the PREVENTION dataset. The obtained results clearly demonstrate the potential of these methodologies  to serve as robust predictors of future lane-changes of surrounding vehicles proving an accuracy higher than $90\%$ in time horizons of between 1-2 seconds.
\end{abstract}


%
\IEEEpeerreviewmaketitle

\section{Introduction}
%
%
%
%
\IEEEPARstart{O}{ne} of the closest and most plausible scenarios in the adoption of the autonomous vehicles is autonomous navigation at SAE L3 (chauffeur) or L4 (autopilot) on highways, both for passenger and freight transport. This is mainly due to the maturity of one of the first driver assistance technologies: Adaptive Cruise Control (ACC) systems. They were introduced in the early 1990s and are present in a wide range of passenger vehicles today \cite{Xiao2010}. ACC systems focus on maintaining a desired speed selected by the driver or maintaining the distance between the car in front and the ego car. Newer versions introduced Stop \& Go functionality. But the steering wheel must be controlled manually (L1). 

The next step in automation were the Traffic Jam Assist (TJA) and the Traffic Jam Chauffeur (TJC) that combines the ACC Stop \& Go and Lane Keeping Assist functions to control the steering wheel, speed, acceleration and braking of the vehicle in traffic jams up to speeds typically below 60 km/h. TJA is usually considered as L2 and TJC as L3 \cite{ERTRAC2019}.

Finally, the most advanced automation systems to date are the Highway Chauffeur (HC) and the Highway Autopilot (HA), which includes the management of complex maneuvers such as deciding to change lanes to overtake, enter a slower lane or even exit the highway. HC is mostly considered as L3 and HA as L4\cite{ERTRAC2019}. 

\begin{figure}[!t]
	\centering	
	\includegraphics[width=0.48\textwidth]{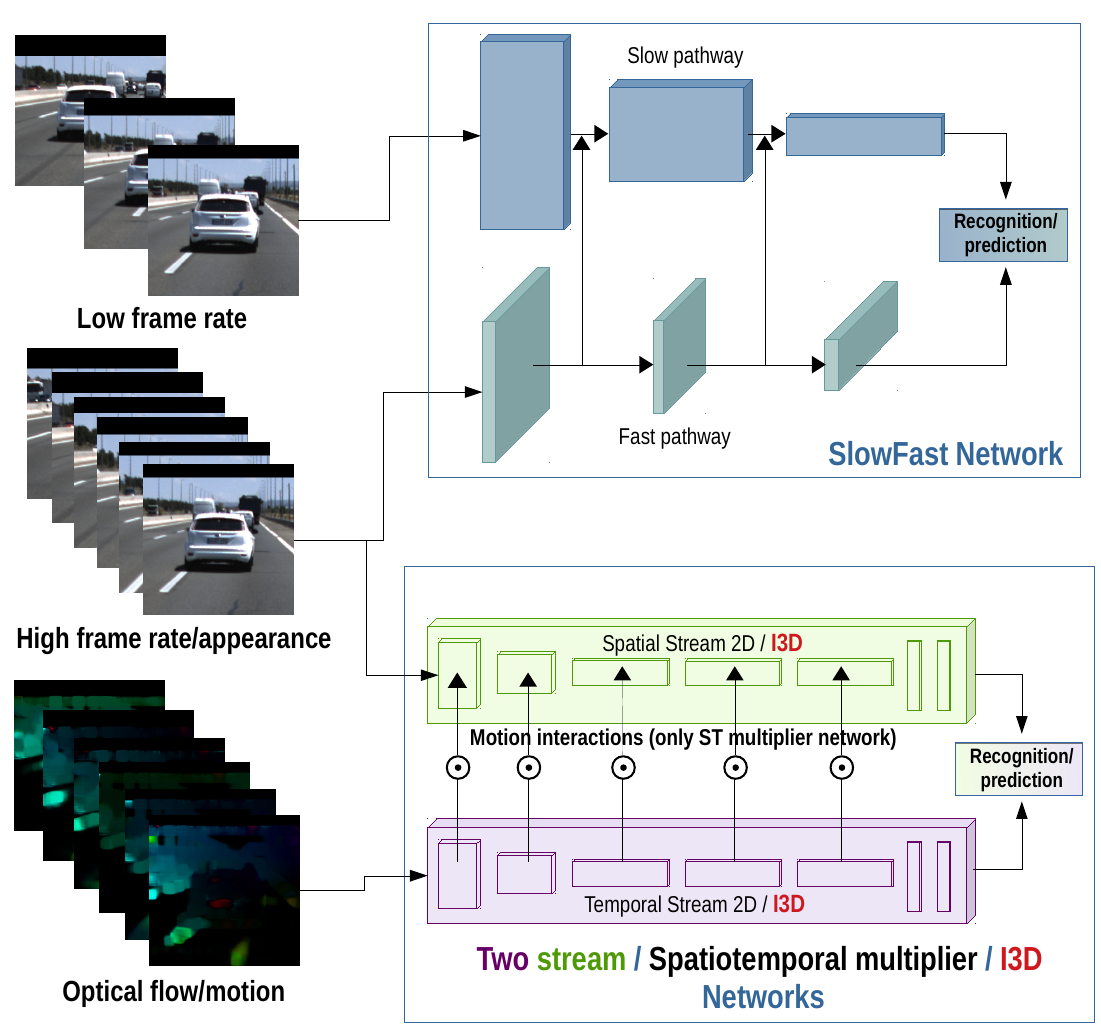}
	\caption{Overview of the proposed video action recognition approaches for lane change recognition and prediction of surrounding vehicles, including Two-Stream Network, Two-Stream Inflated 3D ConvNet, Spatiotemporal Multiplier Network and SlowFast Network.}
	\label{fig:overview}
\end{figure}

In all previous systems, from the simple ACC to the most sophisticated HA, the most critical, and challenging, highway scenarios are the cut-in and cut-out ones, specially for high speeds. In the cut-in scenario, a car from one of the adjacent lanes merges into the lane just in front of the ego car. In the cut-out scenario, a car in front leaves the lane abruptly to avoid a slower vehicle, or even stopped, ahead. Since 2018, the performance of these assistance or chauffeur commercial systems operating under these two critical traffic scenarios is being tested by Euro NCAP \cite{EURONCAP2018}. Although there are abnormal behaviors that can also lead to critical situations on highways, and that are of interest to driving automation systems, such as sudden stops, abnormal trajectories or collisions, the amount of data available is still very limited to develop, validate and certify potential approaches.

An alert driver will typically anticipate cut-in and cut-out maneuvers, even over long distances, using only visual cues, reduce speed accordingly, or even change lanes through the use of the steering wheel. An automated system must also be able to anticipate these situations at an early stage. To do so, it is necessary to endow new automated systems with the ability of predicting the motions of surrounding vehicles, such as lane-keeping and lane-change, and thus improving driving performance significantly in terms of safety, comfort, and even environmental sustainability \cite{Liu2021}, \cite{Llorca2021}.  

The first Lane Departure Warning (L0) or Lane Change/Lane Keeping Assist systems (L1) were designed to detect, or even predict, lane departure of the ego vehicle by combining visual cues (lane markings and lane texture) and vehicle-state information (CAN bus) \cite{McCall2006}. Although there is some ambiguity in the available literature, ego-vehicle lane change detection systems differ considerably from surrounding vehicle lane change detection systems. The requirements for sensors are very different, as are the applicable methodologies. For example, the pixel resolution available to detect the position of vehicles relative to their lane is much lower. Solutions require vehicle detection and tracking. Relative distance and speed measurements require radar or LiDAR type range sensors, and yet, uncertainty of the measurements is much more relevant. Vehicle-state information (e.g., accelerations), can only be accurately obtained via V2V communications \cite{Parra2019}.

To deal with lane-change prediction of surrounding vehicles, in this paper we pose the problem as an action recognition problem using visual information from cameras. The idea behind our proposal is to use the same source of information (visual cues) and the same type of approach (action recognition) that drivers use to anticipate these maneuvers. By using a spatio-temporal model based on image sequences (i.e., continuous visual cues) our approach implicitly includes positional, contextual, and symbolic information, such as turn or brake indicators. Although there are some drivers who do not use them (i.e., a system based solely on their detection would not be effective), in general they are a very valuable source of visual information.

Significant progress has been made in video-based human action recognition and prediction during the last years \cite{Kong2018}. Action recognition and prediction involves managing spatial and temporal information (sequence of images). Among the different methodologies, we focus our efforts in the following two-stream-based approaches (see Figure \ref{fig:overview}):

\begin{itemize}
\item \textbf{Two-Stream Convolutional Networks} \cite{Simonyan2014}: a classical architecture that contains a spatial network and a temporal network (two streams), which are used for modeling static information in still frames and motion information in optical flow images, respectively. 
\item \textbf{Two-Stream Inflated 3D Convolutional Networks (I3D)} \cite{Carreira2017}: an extension of the classical two-stream architecture which expand filters and pooling kernels into 3D, leading to very deep, naturally spatiotemporal classifiers.
\item \textbf{Spatiotemporal Multiplier Networks} \cite{Feichtenhofer2017}: a two-stream architecture that combines appearance and motion pathways and allows interaction between them by injecting cross-stream residual connections. 
\item \textbf{SlowFast Networks} \cite{Feichtenhofer2019}: a two-stream architecture involving a slow pathway that operates at low frame rate to capture spatial semantics and a fast pathway that operates at high frame rate to capture motion at fine temporal resolution. 
\end{itemize}

Although there are other works focused on learning spatiotemporal features for video activity recognition, the selected approaches are a good example of the evolution of the two-stream-based systems, including the first successful proposal \cite{Simonyan2014} and the one ranked first \cite{Feichtenhofer2019} in the AVA Challenge 2019 \cite{gu2017ava}. 
Beyond our previous preliminary work \cite{Llorca2020}, to the best of our knowledge, this is the first proposal using video action recognition approaches to deal with lane-change recognition and prediction of surrounding vehicles for automated vehicles. 

To validate these approaches in this context, we make use of The PREVENTION dataset \cite{prevention_dataset} which  provides  a large  number  of  accurate  and  detailed  annotations  of  vehicles  categories,  trajectories  and  events  (including  left/right lane  changes,  among  others).  More  than  356  minutes,  4M vehicle  detections  and  3K  trajectories  are  available,  with data collected from LIDAR, radar and camera sensors, from surrounding vehicles up to a range of 80 meters.  Contours and bounding boxes are available as raw output detections, as well as a temporary integration of the detections.

The aforementioned architectures are adapted to deal with lane change action recognition and prediction. An extensive evaluation is performed in this paper. The amount of context information needed to model the interactions between different vehicles and other features implicitly included in the appearance, such as the number of lanes or the road curvature, is studied using different sizes for the regions of interest. The ability of the networks to perform action recognition and prediction is assessed using different time horizons and training strategies. The obtained results clearly validate the use of these type of approaches to solve the lane-change prediction problem of surrounding vehicles. 

The remainder of the paper is organized as follows. In Section \ref{sec:soa}, the related work is presented, whereas Section \ref{sec:problem} is an overview of the problem formulation. In Section \ref{sec:approaches} the implementation of the action recognition approaches is described. In Section \ref{sec:experiments} the evaluation metrics, and the performance of the different approaches are assessed. The final conclusions and future work are given in Section \ref{sec:conclusions}.

\section{Related Work}
\label{sec:soa}
Most of the available work on lane-change recognition and prediction focuses on in-vehicle detection. However, as stated before, the nature of the problem is considerably different, so we limit our analysis of lane-change detection of other vehicles, and more specifically, within the context of the highway scenario. Vehicle and lane markings detection and tracking \cite{Sivaraman2013} are necessary conditions. However, it is reasonable to consider them as separated problems that are independent of the maneuver recognition system.

Three levels of analysis will be considered. First, we will review the type of input features used. Second, we will focus on the different types of methodologies. Finally, we will describe the available datasets and their main features. 

\subsection{Input variables}
Most of the works analyzed are based on the use of physical variables that define the relative dynamics of the vehicle with other vehicles and with its environment \cite{Kasper2012}, \cite{Graf2013}, \cite{Schlechtriemen2014}, \cite{Liu2014}, \cite{Schlechtriemen2015}, \cite{Yoon2016}, \cite{Bahram2016},\cite{Izquierdo2017}, \cite{Yao2017}, \cite{Lee2017}, \cite{Deo2018}, \cite{Deo2018b}, \cite{patel2018predicting}, \cite{Li2019e}, \cite{Kruger2019}. Some of these variables are lateral and longitudinal position (distances), velocity, acceleration, timegap, heading angle and yaw rate. These variables are usually obtained and processed in a multi-modal fashion, by fusing data from onboard sensors such as cameras and range sensors (radar and/or LiDAR). Errors and uncertainties in the estimation of these variables from the raw data lead to additional limitations. We can expect reasonably accuracy when measuring the position and relative velocities of other vehicles using onboard sensors. However, it is unrealistic to handle accurate measurements of variables such as lateral and longitudinal accelerations, yaw angle, or yaw rate. As an example, we refer to \cite{Llorca2016} to see the intrinsic difficulty of obtaining accurate speed measurements from static cameras. Sensor uncertainties are intrinsically modeled in some approaches \cite{Kasper2012}, \cite{Li2019e}, but even so, we cannot expect them not to affect predictions.  In some cases it is assumed that these variables will be available via V2V communications \cite{Liu2014}, but this scenario requires a 100\% penetration rate, and in that case, predicting the intentions of other vehicles would be unnecessary as the vehicles could transmit their intentions. In addition, V2V communications pose a number of additional problems to consider \cite{Parra2017}. In any case, we are still far from this scenario.

Context cues are also introduced, including road-level features such as the curvature and speedlimit \cite{Schlechtriemen2014}, \cite{Schlechtriemen2015}, \cite{Li2019e}, distance to the next highway junction \cite{Bahram2016}, number of lanes \cite{patel2018predicting}, etc., as well as lane-level features such as type of lane marking or the distance to lane end \cite{Bahram2016}. These variables are inferred and processed from camera sensors, localization systems and enhanced digital maps, and are also subject to errors and uncertainties that will affect detection and prediction performance.

The number of proposals making use of appearance features to perform lane change recognition or prediction is surprisingly low (excluding vehicle and lane markings detection which are common features in all approaches), especially considering that human drivers do not use the physical variables mentioned above to anticipate lane changes from other vehicles but visual cues. In \cite{Li2019itsc} the position of the vehicle bounding box in the image (in pixels) is used, but no appearance features are extracted. This approach is very sensitive to camera position, orientation and settings. In \cite{Li2019e}, two variables manually selected from the appearance, i.e., state of turn indicators and state of brake indicators, are used. Likewise, these variables are obtained from a specific detection system that involves errors and uncertainties that will affect later stages. But the main limitation of systems that explicitly seek to detect turn indicators is that in many cases drivers do not make use of them when changing lanes. In our previous works \cite{Izquierdo2019}, \cite{Izquierdo2021} regions of interest (ROIs) are generated for each vehicle detection, including local information around the vehicle, and appearance features are extracted using a GoogLeNet pre-trained on ImageNet. Using the raw image data (appearance) as input to the lane change detection and prediction system is challenging, but has the benefit of not requiring intermediate detection steps that can introduce additional errors and uncertainties.

\subsection{Methodologies}
As suggested by \cite{Lefevre2014} vehicle motion modeling and prediction approaches can be classified into three different levels: physical-based, where predictions only depend on the laws of physics, maneuver-based, where the future motion of a vehicle depends on the driver maneuver, and  intention-aware, where predictions take into consideration inter-dependencies between vehicles. Note that, as a chicken-egg problem, on the one hand, lane-change recognition can be addressed using the trajectory estimated by any of the motion models \cite{Messaoud2021}, and on the other hand, the prediction of the trajectories of surrounding vehicles can be estimated more accurately if the lane-change intention recognition is available. 
 
Some proposals are intention-aware in their nature. For example, by using graphical models such as Bayesian Networks \cite{Kasper2012}, \cite{Bahram2016}, \cite{Li2019e} or Structural Recurrent Neural Networks \cite{patel2018predicting}, or by using convolutional social pooling in an LSTM encoder-decoder architecture \cite{Deo2018b}. However, in most cases, inter-dependencies between vehicles are modeled by extracting relative physical features (distances, velocities or time-gaps) \cite{Schlechtriemen2014}, \cite{Schlechtriemen2015}, \cite{Yao2017}, \cite{Deo2018} or by generating compact representations that encode the relative positions of all vehicles on the scene \cite{Lee2017}, \cite{Izquierdo2019}, \cite{Izquierdo2021}. A considerable number of previous works do not take into consideration the interaction between vehicles \cite{Graf2013}, \cite{Liu2014}, \cite{Yoon2016}, \cite{Li2019itsc}, \cite{Kruger2019}. 

Many approaches to lane-change recognition and prediction address the problem using generative-based solutions, including Na\"{i}ve Bayes Classifiers \cite{Schlechtriemen2014}, Bayesian Networks \cite{Kasper2012}, \cite{Bahram2016}, \cite{Li2019e}, and Hidden Markov Models \cite{Liu2014}. Others make use of discriminative solutions such as case-based reasoning \cite{Graf2013}, Random Decision Forest \cite{Schlechtriemen2015}, traditional Neural Networks \cite{Yoon2016}, \cite{Izquierdo2017}, Support Vector Machines \cite{Izquierdo2017}, \cite{Yao2017}, \cite{Li2019itsc}, Gaussian Process Neural Networks \cite{Kruger2019}, and feedforward Convolutional Neural Networks \cite{Lee2017}, \cite{Izquierdo2019}, \cite{Izquierdo2021}. Finally, some other approaches are based on the use of Recurrent Networks including vanilla LSTM \cite{Izquierdo2019} and LSTM encoder-decoder \cite{Deo2018} and multi-modal \cite{Deo2018b} architectures. Consequently, two-stream architectures have not been proposed so far to perform lane change detection and prediction.

\subsection{Datasets}



In order to train learning-based approaches and validate the quality of the proposed solutions, available datasets play a fundamental role. Two type of recording setups are usually proposed depending on the location of the sensors. First, we have datasets captured from the infrastructure using cameras installed on buildings, such as NGSIM HW101 \cite{Ngsim101dataset} or NGSIM I-80 \cite{NgsimI80dataset} datasets, or cameras on-board drones, such as HighD \cite{highddataset}, inD \cite{inddataset} or INTERACTION \cite{interactiondataset} datasets. Although these datasets are very valuable for understanding and assessing the motion and behavior of vehicles and drivers under different traffic scenarios, they are not fully applicable for on-board detection applications. 

Second, other datasets provide road data with sensors on-board vehicles. In this line, the PKU dataset \cite{pkudataset} was released in 2017 by Peking University and the PSA Group, containing 170 minutes of data gathered using a vehicle equipped with 4 2D-LiDARs covering a region of 40 meters around the vehicle. It does not contain information regarding the road lane markings, the number of road lanes, or the relative positioning of the ego-vehicle. In 2018, the ApolloScape dataset \cite{apolloscapedataset} was released by Baidu Research, containing data obtained in urban environments from 4 cameras and 2 Laser scanners using a vehicle driving at 30 km/h. It is currently one of the most complete datasets in the state-of-the-art but it does not contain radar data, making detections more sensitive to failure in adverse weather conditions and highway scenarios. In addition, it does not provide labeled tracking information (IDs and tracklets) for all detected objects. In 2019, the PREVENTION dataset \cite{prevention_dataset} was released containing data from 3 radars, 2 cameras and 1 LiDAR, covering a range of up to 80 meters around the ego-vehicle (up to 200 meters in the frontal area). Road lane markings are included and the final position of the vehicles is provided by fusing data from the three type of sensors. 

\begin{figure*}[!t]
	\centering	
	\includegraphics[width=0.97\textwidth]{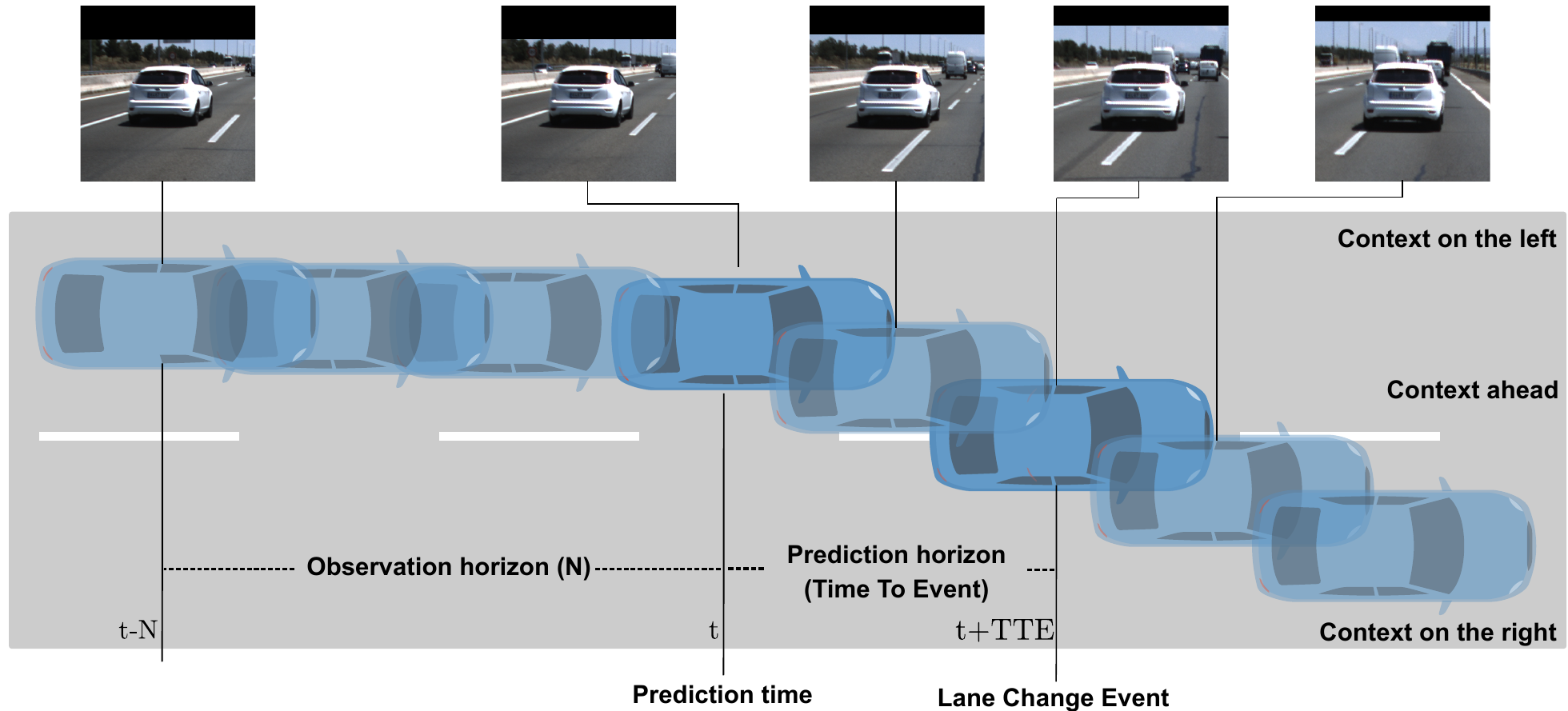}
	\caption{Problem formulation: observation horizon (N), and time to event (TTE). The lane change event is labeled as the frame where the  middle of the rear bumper is located just over the lane markings. This is the criterion established in PREVENTION dataset \cite{prevention_dataset}. }
	\label{fig:problem}
\end{figure*}

\section{Problem Formulation}
\label{sec:problem}
We define lane change prediction as a multi-classification problem in which the goal is to determine whether a vehicle $i$ will make a left or right lane-change (LLC, RLC) or remain in its lane (no lane change) given the observed context up to some time $N$. As can be seen in Figure \ref{fig:problem}, the lane-change event is defined as the time when the center of the rear bumper is just above the lane markings. Therefore, cases with small lateral displacements, lateral oscillations, or aborted lane change maneuvers (including unsafe or aggressive behaviors) are contained in the no lane change class (NLC). Although these are difficult cases that often result in false positives, false lane change detection and prediction would not be as critical for the context of predictive driving automation systems as they can anticipate dangerous situations in which the safest control actions would be the same as in the case of actual lane changes.

The observation horizon or time window will contain a set of $N$ images that will be stacked according to the activity recognition method used. 

Then, the problem can be posed as a \emph{classification} or \emph{prediction} problem based on the value of the Time to Event ($TTE$), or prediction horizon, as follows:
\begin{itemize}
    \item Lane-change \emph{classification}: when $TTE=0$. That is, the observation horizon contains part of the lane change maneuver itself for the LLC and RLC classes.
    \item Lane-change \emph{prediction}: when $TTE>0$. Depending on the TTE value, the observation horizon will contain more or less information of the actual lane change maneuver for LLC and RLC classes. For very high TTE values the maneuver may not even have started. Still, contextual or symbolic information can help anticipate lane changes in these cases.
\end{itemize}
We will examine the effects of $TTE$ or prediction horizon and observation duration ($N$) on the accuracy of lane-change classification and prediction.

\begin{figure}[!t]
	\centering	
	\includegraphics[width=0.45\textwidth]{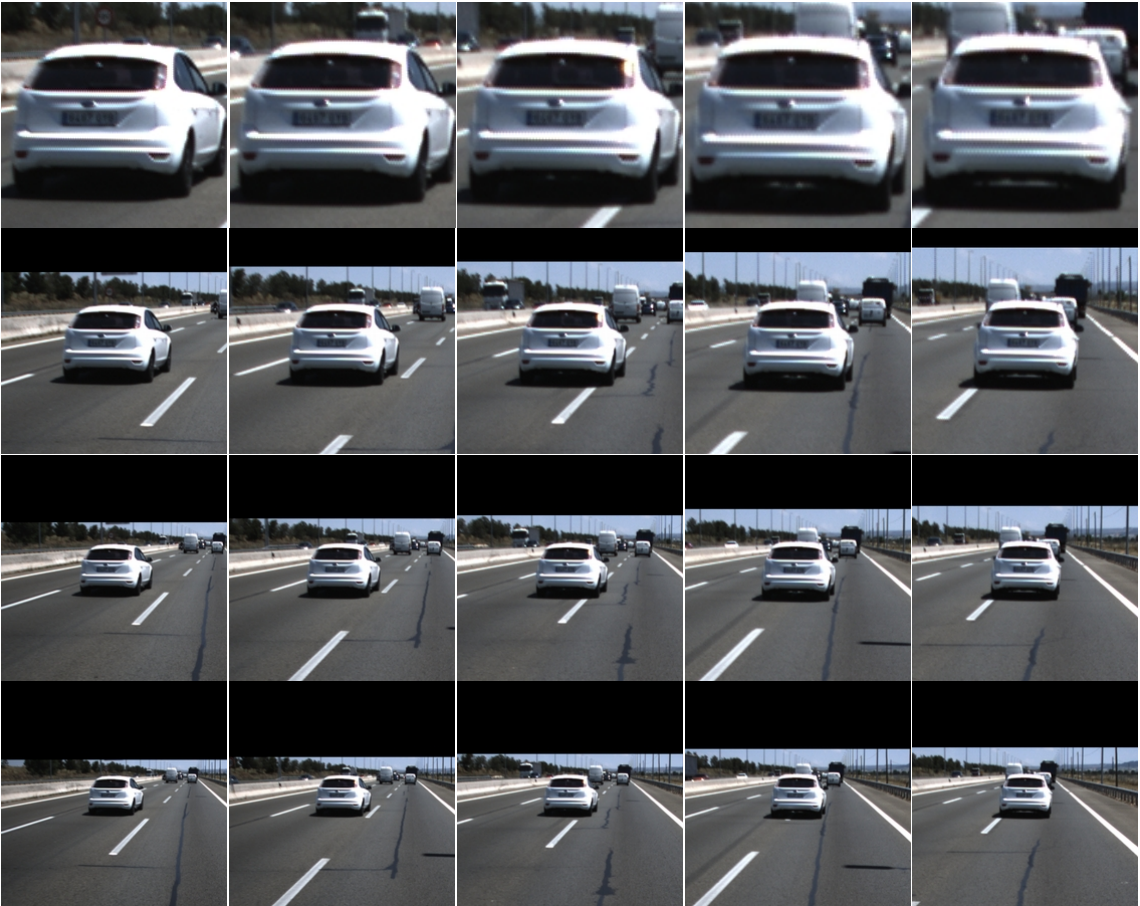}
	\caption{ROI sizes. From upper row to lower row: x1, x2, x3 and x4. The vehicle is always centered. Zero-padding is applied when needed.}
	\label{fig:rois}
\end{figure}

The prediction relies on visual cues that are computed from regions of interest (ROI) extracted from the contour labels provided in the PREVENTION dataset. Four different ROI sizes are considered: $\times 1$, $\times 2$, $\times 3$ and $\times 4$ the size of the square bounding box around the vehicle contour (see Figure \ref{fig:rois}). Zero-padding is used when the ROI exceeds the limits of the image. The size of the ROI modulates the amount of context information being considered in the input data stream. Thus, $\times 1$  mostly contains information related with the vehicle appearance, while $\times 4$ incorporates a large amount of front and side context information. For ROI sizes of $\times 3$ and $\times 4$ the approach can be considered interaction-aware since the image contains information regarding cars in the same or adjacent lanes. Other variables relevant for lane change prediction such as the number of lanes, or road curvature, are implicitly included in the context information.

\begin{figure}[!t]
	\centering	
	\includegraphics[width=0.47\textwidth]{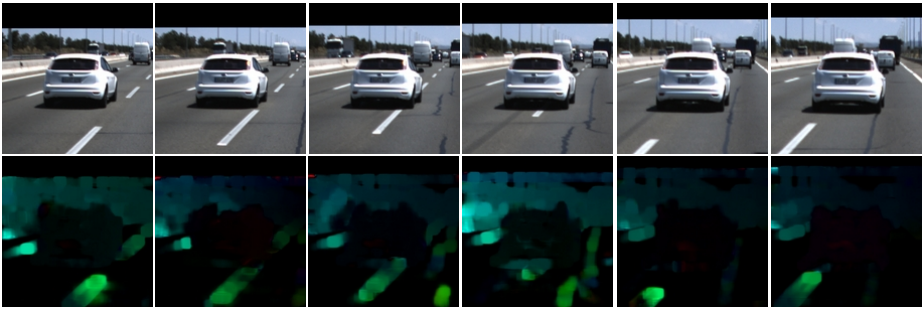}
	\caption{Example of dense optical flow computation.}
	\label{fig:of}
\end{figure}

Since the vehicle is always centered in the ROI, dense optical flow (from the motion stream) should be interpreted as a way of measuring the movement of the context (infrastructure and other vehicles) around the detected vehicle. As shown in Figure \ref{fig:of}, the optical flow is low in the region where the vehicle is, while it is more predominant around it.


\section{Video Activity Recognition \& Prediction}
\label{sec:approaches}
The sequence of stacked images or regions of interests, can naturally be decomposed into spatial and temporal components. The spatial part, in the form of individual region appearance, carries information about the vehicle itself (e.g., light indicators or brake lights) and the context around it (road, lane markings and surrounding vehicles). The temporal part, in the form of motion across frames, conveys the movement of the observer (onboard camera) w.r.t. to the road, and the surrounding vehicles. In order to handle a canonical view for the motion stream, all the regions are generated around the contour of the vehicle so the vehicle is always centered in the region of interest (the size will vary depending on the relative distance w.r.t. the ego vehicle). We consider four video activity recognition approaches: Disjoint Two-Stream Convolutional Networks (TS) \cite{Simonyan2014}, Two-Stream Inflated 3D Convolutional Networks (I3D) \cite{Carreira2017},  Spatiotemporal Multiplier Networks (STM) \cite{Feichtenhofer2017} and SlowFast Networks (SF) \cite{Feichtenhofer2019}.

\begin{figure*}[!t]
	\centering	
	\includegraphics[width=0.85\textwidth]{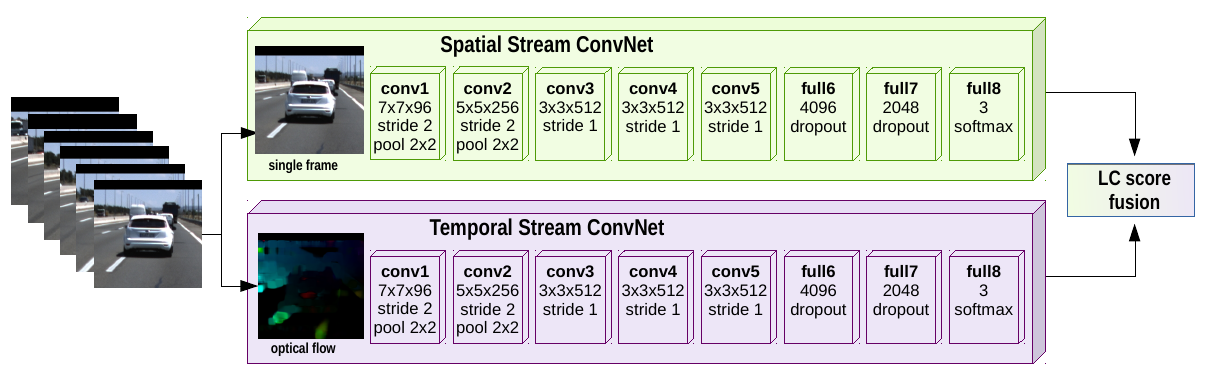}
	\caption{Disjoint two-stream architecture for lane change classification and prediction.}
	\label{fig:disjoinConvNet}
\end{figure*}

\subsection{Disjoint Two-Stream Convolutional Networks}
A two-stream ConvNet architecture which incorporates and fuses spatial and temporal information is defined. The structure of the ConvNets for both streams is the same, including 5 convolutional layers and 3 fully connected layers, with the parameters depicted in Figure \ref{fig:disjoinConvNet}. The last fully connected layer is defined with 3 outputs regarding the three classes defined: left lane change (LLC), right lane change (RLC), and no lane change (NLC).

The dense optical flow is computed using polynomial expansion \cite{Farneback2003}. The spatial stream ConvNet is pre-trained using ImageNet and the temporal ConvNet using multi-task learning using UCF-101 and HMDB-51. All hidden layers use the rectification (ReLU) activation function. Max-pooling is performed over $3\times 3$ spatial windows with stride 2. 

\subsection{Two-Stream Inflated 3D Convolutional Networks}
\begin{figure}[!t]
	\centering	
	\includegraphics[width=0.3\textwidth]{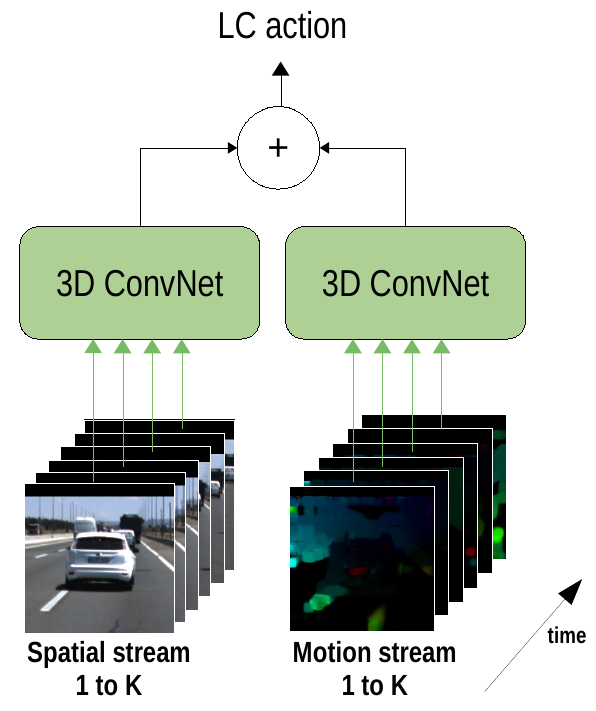}
	\caption{Two-stream inflated 3D ConvNet for lane change classification and prediction.}
	\label{fig:i3d}
\end{figure}

The natural approach to deal with video modeling is to use 3D convolutional neural networks. These are like standard convolutional networks, but with spatio-temporal filters that generate a hierarchical representation of spatio-temporal data. These are more complex architectures with a higher number of parameters that cannot easily benefit of pre-training strategies. We adopt the approach presented in \cite{Carreira2017} which starts with a 2D architecture and inflates all the filters and pooling kernels endowing them with an additional temporal dimension. Each 3D network is implemented with 8 convolutional layers, 5 pooling layers and 2 fully connected layers at the top. Batch normalization is applied after all convolutional and fully connected layers. The 3D filters are bootstrapped from pre-trained ImageNet models by repeatedly copying an image into a video sequence. A two-stream configuration is used (see Fig. \ref{fig:i3d}), learning temporal patterns from the appearance stream, but enhancing its performance by including the motion stream. The inputs to the model are short 16-frames sequences.

\subsection{Spatiotemporal Multiplier Networks}

The original two-stream architecture only allows the two processing streams (spatial and motion) to interact via late fusion of their respective softmax predictions. This way, the architecture does not support the learning of truly spatiotemporal features, since the loss of both streams is backpropagated independently without any type of interaction. Learning spatiotemporal features requires the appearance and motion paths to interact earlier on during the forward pass. This interaction can be relevant for the classification and prediction of lane change maneuvers that have similar appearance or motion patterns and can only be inferred by the combination of two (e.g., vehicles that do not change lanes but have their turn indicators on). To address this limitation, it is possible to inject cross-stream residual connections using Residual Networks (ResNets) \cite{He2016} as the general architecture for the spatial and the temporal streams.

In \cite{Feichtenhofer2017}, different cross-stream connections were studied, including two types of connections (direct or into residual units), two fusion functions (additive or multiplicative), and different streams directions (unidirectional from the motion into the appearance, conversely and bidirectional), being the multiplicative residual connection from the motion path into the appearance stream the one providing the superior performance. 

\begin{figure}[!t]
	\centering	
	\includegraphics[width=0.3\textwidth]{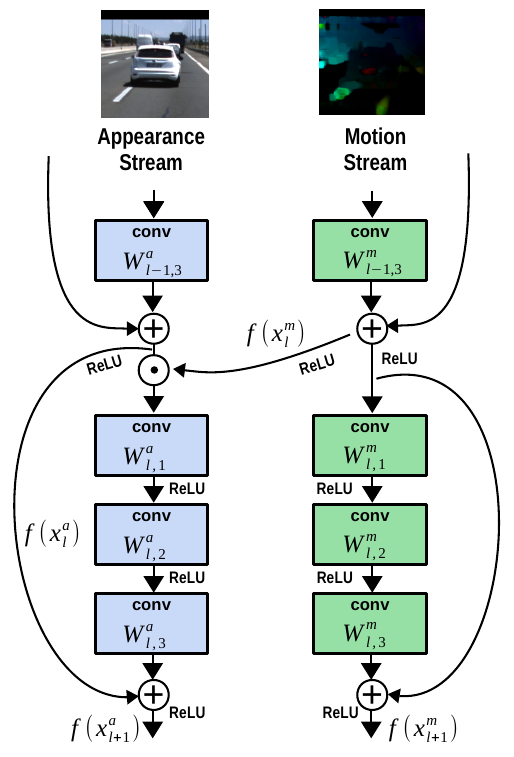}
	\caption{Multiplicative residual gating from the motion stream to the appearance stream.}
	\label{fig:stx}
\end{figure}

As can be observed in Figure \ref{fig:stx}, the multiplicative interaction can be formulated as:

\begin{equation}
\hat{x}^a_{l+1}=f(x_l^a)+\mathcal{F}\Big(x_l^a\odot f(x_l^m),W_l^a \Big)
\end{equation} 
where $x_l^a$ and $x_l^m$ are the inputs of the $l$-th layers of the appearance and motion paths respectively, while $W_l^a$ represents the weights of the $l$-th layer residual unit in the appearance stream and $\odot$ corresponds to elementwise multiplication. 
 
Better temporal support is also provided by injecting 1D temporal convolutions layers into the network \cite{Feichtenhofer2017}. ResNet50 model is used for both streams, including batch normalization and ReLU activation function after each convolutional block. 

\subsection{SlowFast Networks}
\begin{figure}[!t]
	\centering	
	\includegraphics[width=0.45\textwidth]{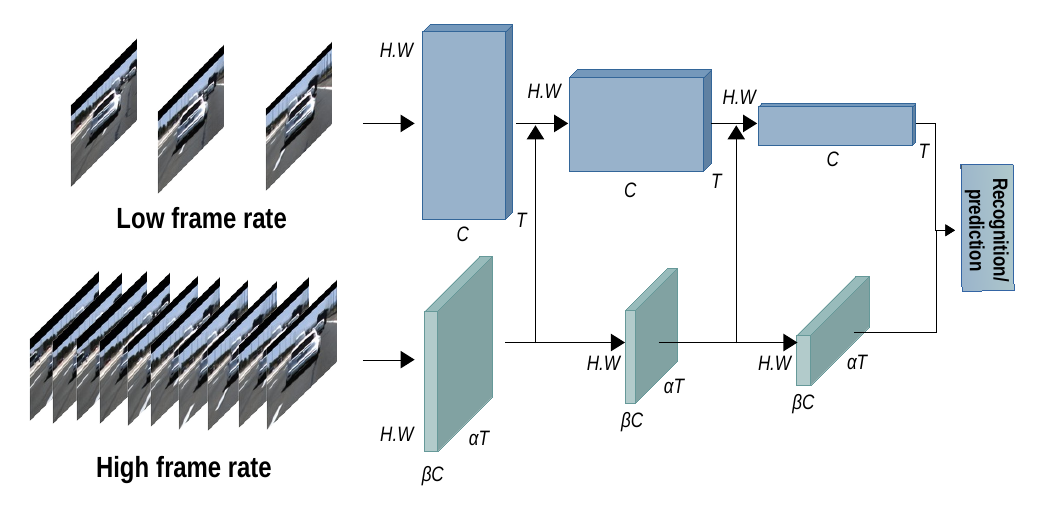}
	\caption{SlowFast network for lane change recognition and prediction. The fast stream is lightweight by using a fraction $\beta=1/8$ of channels.}
	\label{fig:sfn}
\end{figure}

One of the most successful video action recognition approaches is the so called SlowFast network \cite{Feichtenhofer2019}. It can be considered as a two-stream approach, although motion pathway is not directly used. Instead, one stream (slow) is designed to capture semantic information given by a few sparse images operating at low frame rates and slow refreshing speed, and a second stream (fast) is responsible for capturing rapidly changing motion by operating at high temporal resolution and fast refreshing speed. The two pathways are fused by lateral connections. The temporal stride used in the Slow pathway is $\tau=16$ and the frame rate ratio between the Fast and Slow streams is $\alpha=8$. The ratio of channels of the Slow stream with respect to the Fast one is defined as $\beta=1/8$ (see Fig. \ref{fig:sfn}). The network is defined with one convolutional layer , five residual blocks and one fully connected layer adapted to the number of classes as in \cite{Feichtenhofer2019}. Since optical flow is not computed, the architecture can be learned end-to-end from the raw data. 

\section{Experiments}
\label{sec:experiments}

\subsection{Dataset description}
Table \ref{table:dataset} summarizes the details of the dataset. The input size for both streams is $112\times112$. The $85\%$ of the samples are used for training and the remaining $15\%$ for validation.  

\begin{table}[h!]
\centering
\caption{Main stats of the dataset. NLC/LLC/RLC: no/left/right lane-change.}
\begin{tabular}{|c|c|c|c|} 
  \hline
 & NLC & LLC & RLC \\
 \hline
 \# of sequences & 3110 & 342 & 438 \\ 
 avg. \# of frames & 50.9 & 96.8 & 80.1 \\ 
 \hline
\end{tabular}
\label{table:dataset}
\end{table}

\subsection{Evaluation parameters and models}
The following parameters have been evaluated during the experiments:

\begin{itemize}
\item ROI sizes: x1, x2, x3 and x4.
\item Observation horizon: 20 frames (2 seconds), 30 frames (3 seconds) and 40 frames (4 seconds).
\item Time-to-event (prediction horizon): 0 (no prediction), 10 (1 second) and 20 (2 seconds).
\end{itemize}

The evaluated video recognition models are the Disjoint Two-Stream ConvNet (\textbf{Disjoint}), the Two-Stream Inflated 3D ConvNet (\textbf{I3D}), the Spatiotemporal Multiplier ConvNet (\textbf{ST}), and the SlowFast ConvNet (\textbf{SF}). A basic model which implements the appearance stream of the Disjoint architecture (upper pathway in Fig. \ref{fig:disjoinConvNet}) is used as the baseline (\textbf{Baseline}). In all cases, the specific architecture of the models and the hyper-parameters used for training are those reported as optimal by the authors.

\subsection{Metrics}
As a multi-class problem (with 3 classes), we use the categorical entropy loss function for the training. For evaluating the results, we consider the accuracy as the main variable to assess the performance of the two evaluated methods and the corresponding parameters, i.e., the number of true positives for the three classes divided by the total number of samples (arithmetic mean of precision for all classes). In addition, we evaluate precision and recall for all classes in confusion matrices. 

\subsection{Lane-change classification results}
In Table \ref{table:1} we depict the lane-change classification (i.e., with $TTE=0$) accuracy of all action recognition approaches over the validation set. 

Regarding the ROI sizes we can state the following conclusions. By using just the ROI fitted to the bounding box, the results are surprisingly reasonable, considering that almost no context and interaction are available. In general, the higher the ROI size, the better the accuracy (with the exception of the I3D model), although adding more context from $x3$ to $x4$ decreases the performance for most cases and observation horizons. This can be explained by the fact that the observation horizons already incorporate context into the spatial, motion and slow streams, so using a higher ROI is not reflected in a better performance. 

The effect of the observation horizon depends on the model. For example, the Disjoint and the Spatiotemporal models yield the best classification performance with the longest observation horizon (4 seconds). However, the I3D and the SlowFast architectures have a higher accuracy with the shortest observation horizon (2 seconds). For the classification task, the last frames are the most informative, and these models (I3D and SlowFast) seem to take better advantage of this information without the need for a larger observation horizon. 

The best classification results, $90.98\%$, are provided by the SlowFast model with a ROI size of $x3$ and an observation horizon of 2 seconds, followed by the Spatiotemporal Multiplier Network, $90.30\%$ with a ROI size of $x3$ and an observation horizon of 4 seconds. 


\begin{table}[h!]
\centering
\caption{Lane-change Classification ($TTE=0$) Accuracy ($\%$).}
\begin{tabular}{|c|c|c|c|c|c|} 
 \hline
 & & \multicolumn{4}{c|}{ROI size} \\
 \hline
 Method & Obs. Horizon & x1 & x2 & x3 & x4\\
 \hline
 \multirow{3}{*}{Baseline} & 20 & 83.41 & 83.25 & \textbf{85.35} & 84.06 \\ 
                           & 30 & 81.96 & 83.25 & 82.61 & 85.19 \\
						   & 40 & 81.80 & 82.45 & 81.32 & 81.80 \\ 
 \hline
 \multirow{3}{*}{Disjoint} 	& 20 & 83.22 & 86.18 & 86.26 & 87.43 \\ 
  							& 30 & 83.55 & 86.69 & 86.84 & 86.68 \\
							& 40 & 84.97 & 87.69 & \textbf{89.46} & 88.79 \\ 
 \hline
 \multirow{3}{*}{I3D} 	& 20 & 82.45 & \textbf{86.47} & 85.99 & 85.67 \\ 
  							& 30 & 82.13 & 83.74 & 83.90 & 84.06 \\
							& 40 & 82.13 & 83.09 & 81.80 & 82.29 \\ 
 \hline
 \multirow{3}{*}{ST} & 20 & 83.39 & 85.03 & 86.51 & 86.16 \\ 
  					 & 30 & 84.38 & 84.70 & 85.36 & 84.73 \\
					 & 40 &  86.02 & 87.83 & \textbf{90.30} & 89.64 \\ 
 \hline
 \multirow{3}{*}{SF} & 20 & 88.89 & 89.69 & \textbf{90.98} & 89.37 \\ 
  					 & 30 & 88.57 & 89.53 & 88.24 & 89.69 \\
					 & 40 &  86.96 & 89.05 & 89.53 & 90.34 \\ 				 
 \hline
 \end{tabular}
\label{table:1}
\end{table}

\subsection{Lane-change prediction results}
The ability of all methodologies to predict the future lane-change maneuverer of target vehicles is evaluated using an observation horizon of 20 frames (2 seconds) and prediction horizons of 10 and 20 frames (1 and 2 seconds respectively). The results for all approaches are depicted in Table \ref{table:2}. 

Starting from the baseline, and with the exception of the I3D model, it is remarkable to see that predictions are better for longer prediction horizons, i.e., the obtained accuracy for $TTE=20$ frames is generally higher than for a $TTE=10$ images. This can be explained, in part, by the complexity of the models that better generalize with a more complex objective to learn. This effect is particularly visible with the Disjoint and ST models, where the accuracy is approximately $5\%$ higher when predicting 2 seconds ahead than 1 second ahead. 

Concerning the ROI size we can state that the larger the ROI the better the prediction accuracy, with no saturation effect from $x3$ to $x4$. The best performance when predicting lane-changes 1 second before they occur is obtained with the SlowFast model with a ROI size of $x2$, yielding an accuracy of $88.96\%$. For the larger prediction horizon, 2 seconds, the best model is the Spatiotemporal Multiplier network, which provides an accuracy of $91.94\%$ with a ROI size of $x4$. Note that, the results of the SlowFast model for this case are inconclusive due to GPU memory problems. Whereas the mini-batch size for the other models was 32, the maximum size allowed with the SlowFast model was only 8. It is very likely that without this limitation, the SlowFast model would have provided even better results.


\begin{table}[h!]
\centering
\caption{Lane-change prediction accuracy ($\%$). Observation horizon = 20 frames (2 seconds).}
  \begin{threeparttable}
\begin{tabular}{|c|c|c|c|c|c|}
 \hline
 & & \multicolumn{4}{c|}{ROI size} \\
 \hline
 Method & TTE & x1 & x2 & x3 & x4\\
 \hline
 \multirow{2}{*}{Baseline} & 10 & 82.63 & 82.95 & 83.44 & 82.79 \\
						   & 20 & 82.00 & 81.67 & 82.79 & \textbf{83.61} \\ 
 \hline
 \multirow{2}{*}{Disjoint} & 10 & 84.05 & 84.54 & 85.20 & 85.36 \\
						   & 20 & 85.20 & 88.82 & \textbf{91.02} & 90.92 \\ 
 \hline
 \multirow{2}{*}{I3D} & 10 & 81.33 & 83.28 & 83.60 & 83.60 \\
				      & 20 & 81.01 & 81.67 & \textbf{83.93} & 83.61 \\ 						   
 \hline
 \multirow{2}{*}{ST} & 10 & 84.70 & 85.69 & 85.20 & 86.51 \\
  					 & 20 & 86.84 & 90.30 & 91.45 & \textbf{91.94} \\ 
 \hline
 \multirow{2}{*}{SF} & 10 & 85.23 & \textbf{88.96} & 88.64 & 87.99 \\
  					 & 20\tnote{*} & 85.27 & 83.31 & 83.61 & 83.61 \\   					 
\hline 
\end{tabular}
\begin{tablenotes}
  \item[*] Inconclusive results due to GPU memory limitations. 
\end{tablenotes}
\end{threeparttable}
\label{table:2}
\end{table}

If we analyze the results further, we find that the predictions are closely linked to the number of samples available for each class. In fact, as shown in Table \ref{table:dataset}, we have an unbalanced dataset which clearly affects the results. In Table \ref{table:confmat1} we depict the confusion matrix for the best model (Spatiotemporal Multiplier Network) and the best parameters ($x4$, $TTE=20$) including precision and recall. 

\begin{table}[h!]
\centering
\caption{Spatiotemporal Multiplier Network Confusion Matrix, OH=20, TTE=20, x4} 
\begin{tabular}{c|ccc|c} 
  & \multicolumn{3}{c}{\textbf{Target class}} & \\ 
 \textbf{Output class} & \textbf{NLC} & \textbf{LLC} & \textbf{RLC} & \textbf{Precision} \\
 \hline
\textbf{NLC} & 476 & 5 & 6 & $97.7\%$ \\
\textbf{LLC} & 8 & 33 & 11 & $63.5\%$ \\
\textbf{RLC} & 11 & 8 & 50 & $72.5\%$ \\
 \hline
 \textbf{Recall} & $96.2\%$ & $71.7\%$ & $74.6\%$ & $91.9\%$\\
\end{tabular}
\label{table:confmat1}
\end{table}

As can be observed, the highest precision/recall ratio is obtained for the NLC class which represents almost the $80\%$ of the samples. This correlation between the accuracy and the number of samples is also observed between the LLC and RLC classes, with a better precision/recall ratio for RLC class which contains $28\%$ more samples than LLC. Some of the false positives for the LLC and RLC classes are due to instances of small lateral displacements, or aborted lane change maneuvers. However, the number of samples for these cases is not significant enough to draw further conclusions.

In any case, these are ones of the first prediction results so far using the PREVENTION dataset and the ability of the proposed two-stream multiplier network to predict lane-changes 2 seconds of anticipation is remarkable compared to the ability of humans trying to perform the same task (see \cite{Quintanar2020} and \cite{Izquierdo2021} for more details on human performance).

\section{Conclusions and Future Work}
\label{sec:conclusions}
In this work, four video action recognition approaches have been adapted, trained and evaluated to deal with lane-change classification and prediction of target vehicles in highway scenarios using the labeled images and sequences available in the PREVENTION dataset. The anticipation of lane-changes is devised as an action recognition problem using visual cues from front view cameras, which is the same approach used by human drivers to predict these maneuvers. The Disjoint Two-Stream ConvNets (Disjoint), the Two-Stream Inflated 3D ConvNets (I3D), and the Spatiotemporal Multiplier ConvNets (ST) are based on two different pathways obtained from the same sequence of images: a spatial stream  in the form of individual region appearance, and a motion stream in the form of dense optical flow across frames. The SlowFast ConvNet (SF) is based on two different pathways obtained from the appearance, but taken with two different sampling frequencies (one fast and one slow). 

The influence of the context has been evaluated by using different ROI sizes, being the larger regions ($x3$ and $x4$ the original size of the vehicle) the ones providing the better classification and prediction results. For lane-change recognition, different observation horizons have been tested. Whereas the Disjoint and the ST models yield the best results with the longest observation horizon, the SF network provides the best recognition performance ($91\%$) using visual cues from the shortest observation horizon evaluated, 2 seconds. The I3D model slightly outperforms the baseline, but its classification performance is lower than the rest. 

The ability of most of these models (Disjoint, ST and SF) to predict lane-changes at $t+TTE$ is even better than their ability to classify them at $t$, which is a remarkable feature that can be partially explained by the high complexity of the models that provide better generalization with a more complex
objective to learn. The best prediction results are obtained with the ST model with ROI size of $x4$ and observation horizon of 2 seconds, anticipating lane-changes 2 seconds earlier with an accuracy of $63.5\%$ for left lane-changes, of $72.5\%$ for right lane-changes and of $97.7\%$ for no lane-change.

The presented video action recognition approaches are highly data-dependent, and the number of publicly available datasets is limited. As future works we plan to mitigate the problem with imbalanced classes of the dataset by applying resampling and data augmentation techniques, including generative adversarial networks (GANs) \cite{Wu2019}. Other available and new datasets will be considered to alleviate this problem, to reduce potential bias and to analyze the generalization capabilities of action recognition methods. In cases where synchronized data from multiple sensors are available, the study of data fusion techniques will be addressed to overcome possible limitations of the vision-based method. GPUs memory limitations with models such as SlowFast will be also addressed. Other interesting topics to explore are the impact of lighting and weather conditions on system performance, and the potential ability of the system to explicitly detect and differentiate turn signals and brake lights (e.g., to assess whether or not a driver is correctly signaling an intention to change lanes). Finally, a real-time version will be devised to be tested in on-line real scenarios with our vehicle platforms.

\section*{Acknowledgment}
This work was supported in part by Spanish Ministry of Science, Innovation and Universities (Salvador de Madariaga Mobility Grant PRX18/00155 and Research Grants DPI2017-90035-R and PID2020-114924RB-I00) in part by the Community Region of Madrid (Research Grant 2018/EMT-4362 SEGVAUTO 4.0-CM) and in part by the Air Force Office of Scientific Research USA (FA9550-18-1-0054) the Canada Research Chairs Program (950-231659) and the Natural Sciences and Engineering Research Council of Canada (RGPIN-2016-05352).

\ifCLASSOPTIONcaptionsoff
  \newpage
\fi



\bibliographystyle{IEEEtran}
\bibliography{IEEEabrv,references}

%

%

\begin{IEEEbiography}[{\includegraphics[width=1in,height=1.25in,clip,keepaspectratio]{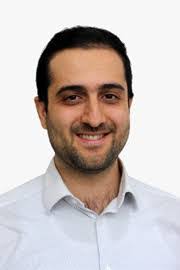}}]{Mahdi Biparva}
received the Ph.D degree in computer science from the York University in 2019. He is currently a Research \& Development Engineering at the Sunnybrook Research Institute. He was Teaching Assistant with York University between 2013-2019 and Research Assistant with the Lab. for Active and Attentive Vision, York Univ. His current research interests are focused on computer vision and deep learning approaches for different tasks in healthcare.
\end{IEEEbiography}

\begin{IEEEbiography}[{\includegraphics[width=1in,height=1.25in,clip,keepaspectratio]{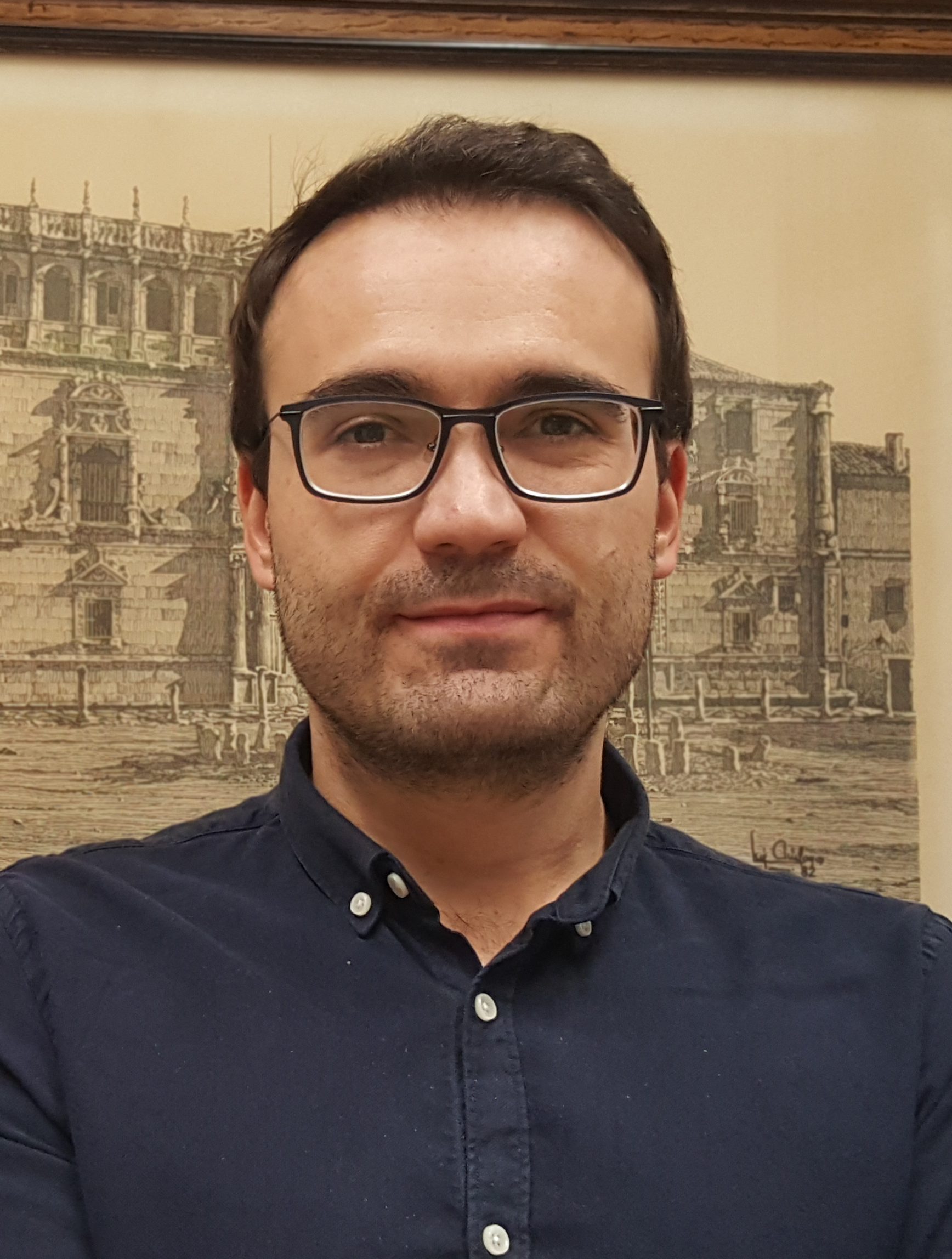}}]{David Fernández Llorca}
received the Ph.D degree in telecommunication engineering from the University of Alcalá (UAH) in 2008. Since November 2020 he is Scientific Officer at the European Commission - Joint Research Center, collaborating in the HUMAINT project. He is Full Professor (special leave) with UAH and co-head of the Intelligent Vehicles and Traffic Technologies (INVETT) research group. He has authored over 130 publications and more than 10 patents. He received the IEEE ITSS Young Research Award in 2018 and the IEEE ITSS Outstanding Application Award in 2013. He is currently Editor-in-Chief of the IET Intelligent Transport Systems. He was Associate Editor of the IEEE Transactions on ITS (2012-2020) and Journal of Advanced Transportation (2016-2020). He was the Program Chair of the IEEE ITSC 2019. His current research interest includes trustworthy AI for autonomous vehicles, predictive perception for autonomous vehicles, human-vehicle interaction, end-user oriented autonomous vehicles and assistive intelligent transportation systems. 
\end{IEEEbiography}


\begin{IEEEbiography}[{\includegraphics[width=1in,height=1.25in,clip,keepaspectratio]{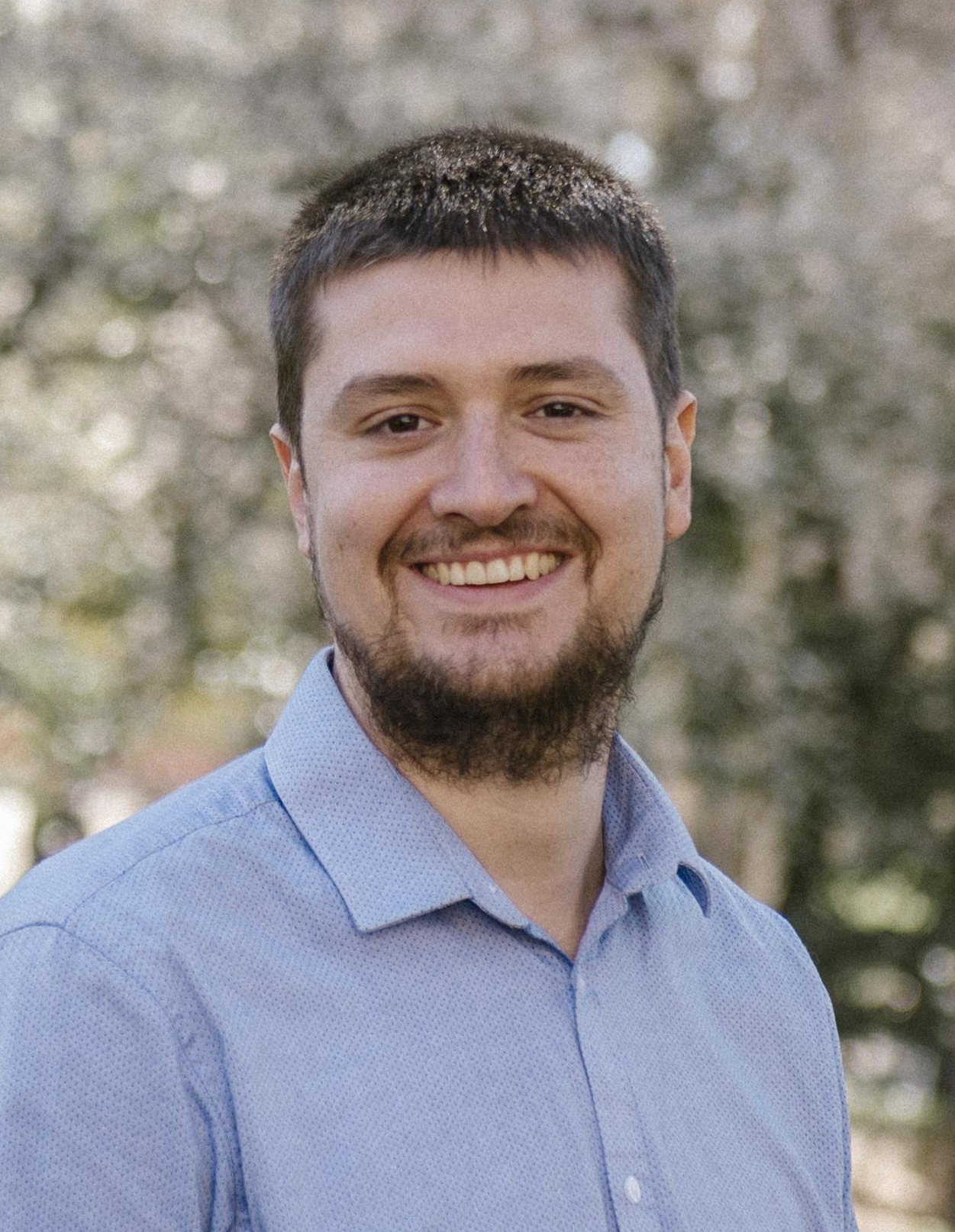}}]{Rubén Izquierdo Gonzalo}
received the M.S. and Ph.D degree in industrial engineering from the University of Alcalá (UAH) in 2018 and 2020, respectively. He is currently a post-doc researcher at the INVETT (Intelligent Vehicles and Traffic Technologies) group. He was the main developer of the DRIVERTIVE team that received the Best Team with Full Automation in the Grand Cooperative Driving Challenge 2016. He received the Social Transfer Council Award UAH in 2018. His research interest includes predictive prediction for autonomous vehicles and cooperative autonomous driving. 
\end{IEEEbiography}

\begin{IEEEbiography}[{\includegraphics[width=1in,height=1.25in,clip,keepaspectratio]{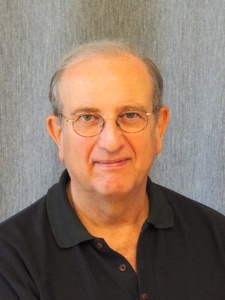}}]{John K. Tsotsos}
is Distinguished Research Professor of Vision Science with York University. He received the Ph.D. degree in computer science from the University of Toronto. He was on faculty in the Department of Computer Science starting in 1980, where he founded the university's Computer Vision Group, which he led for 20 years. He was recruited to York University in 2000 as the Director of the Centre for Vision Research. He is a Fellow of the Royal Society of Canada, has been a CIfAR Fellow, and has several paper prizes and other awards. His current research interest includes comprehensive theory of visual attention in humans. A practical outlet for this theory embodies elements of the theory into the vision systems of mobile robots.

\end{IEEEbiography}




\end{document}